%% file: conference_101719.tex
\documentclass[conference]{IEEEtran}
\IEEEoverridecommandlockouts
\usepackage{cite}
\usepackage{amsmath,amssymb,amsfonts}
\usepackage{algorithmic}
\usepackage[nolist]{acronym}
\usepackage{hyperref}
\usepackage{graphicx}
\usepackage{textcomp}
\usepackage{tabularx}
\usepackage[table]{xcolor}
\usepackage{xcolor}
\definecolor{peachlight}{HTML}{FFE6CC}
\usepackage{tikz}
\newcommand{\colorsquare}[1]{\protect\tikz\protect\node[draw,fill=#1,inner sep=0pt,minimum size=1.5ex]{};}

\def\BibTeX{{\rm B\kern-.05em{\sc i\kern-.025em b}\kern-.08em
    T\kern-.1667em\lower.7ex\hbox{E}\kern-.125emX}}
\begin{document}

\title{Edge AI on Constrained Devices for Binary Sleep-Wake Classification in Dynamic Environments
}


\author{
	\IEEEauthorblockN{Stefan Reitmann\IEEEauthorrefmark{1}, Lena Oden\IEEEauthorrefmark{2}}
\IEEEauthorblockA{\IEEEauthorrefmark{1}\textit{Faculty of Computer Science}, 
\textit{Chemnitz University of Technology}, 
Chemnitz, Germany \\
Email: stefan.reitmann@informatik.tu-chemnitz.de}
\IEEEauthorblockA{\IEEEauthorrefmark{2}\textit{Faculty of Computer Science}, 
	\textit{University of Hagen}, 
	Hagen, Germany \\
	Email: lena.oden@fernuni-hagen.de}
}

\maketitle

\begin{abstract}
This paper presents an Edge AI–based system for detecting sleep and wake states in non-stationary mobile environments using resource-constrained embedded hardware. Conventional approaches relying on accelerometer-based activity metrics are highly susceptible to motion and vibration artifacts and are limited by strict compute and energy budgets of wearable and IoT devices. To address these challenges, a multimodal pipeline is designed and implemented on an ESP32-S3 microcontroller.

The system combines inertial sensing for head movement analysis and visual pose classification. A dual-core architecture with FreeRTOS enables parallel execution of real-time data acquisition and on-device inference. Sleep detection follows a two-stage strategy: low-movement detection over a temporal window, followed by visual validation of poses.

Experimental results show accuracies of 96.5\% for motion-based detection and 89\% for pose classification, yielding robust binary sleep–wake classification. Field tests confirmed feasibility in representative mobile scenarios. The results demonstrate that privacy-preserving, local sleep detection is achievable on edge hardware through careful co-design, while highlighting limitations in sensing intrusiveness, dataset scale, and system integration.
\end{abstract}

\begin{IEEEkeywords}
edge ai, machine learning, embedded systems, microcontroller, sleep detection
\end{IEEEkeywords}

\section{Introduction}
Sleep-wake state detection has become increasingly important across a wide range of applications, including health monitoring, recreational use, and safety-critical systems. Recent advances in wearable and \ac{iot} devices enable continuous acquisition of physiological and motion-related sensor data, providing the foundation for automated sleep analysis. These systems offer the potential for unobtrusive, long-term monitoring outside controlled laboratory environments.

However, in many real-world scenarios, such systems operate under non-ideal measurement conditions. In particular, mobile or dynamic environments - such as vehicles or bicycle trailers . introduce additional motion and vibration that affect sensor readings. These external motion artifacts can overlap with the relevant physiological signals, significantly complicating the reliable discrimination between sleep and wake states.

Another key challenge arises from the limited computational and energy resources of wearable and IoT devices. Since such systems typically rely on low-power embedded hardware, algorithms for state classification must be both robust to noisy sensor data and efficient in their execution. Computationally complex models are often impractical, while sufficient classification accuracy must still be maintained. Against this background, there is a clear need for methods that enable reliable sleep-wake detection under motion-intensive conditions while remaining suitable for deployment on resource-constrained edge devices.

This paper addresses this challenge by developing and prototypically implementing an Edge AI-based system for automatic sleep-wake classification on embedded hardware in mobile environments. The approach focuses on processing sensor data - primarily motion and visual signals - directly on-device, enabling privacy-preserving and energy-efficient operation without reliance on cloud-based computation. The following research questions guide this work:


\begin{enumerate}
    \item How can motion artifacts originating from mobile platforms be mitigated to enable robust sleep detection on edge devices?
    \item Which features and models provide the best trade-off between accuracy and computational efficiency under resource constraints?
    \item To what extent, and in which application scenarios, does sensor fusion improve robustness compared to motion-only approaches?
\end{enumerate}

The system is evaluated experimentally with respect to classification accuracy, computational cost, energy consumption, and robustness to motion artifacts. The results provide insight into the feasibility of reliable sleep–wake detection under realistic operating conditions and identify promising methodological approaches for resource-constrained Edge AI systems.

The source code, trained models, and anonymized intertial datasets are publicly available under an open-source license at \url{https://github.com/Reitmania/esp32_sleepdetection}. The repository includes all necessary implementation details for reproduction and extension of the presented results. Camera data are excluded due to privacy considerations.

\subsection{Background and Related Work}
\label{sec:background}

\subsubsection{Embedded Systems and Edge AI}

Embedded systems are a fundamental component of modern cyber-physical applications, performing specialized computational tasks under strict constraints in processing power, memory, and energy consumption \cite{kopetz_real-time_2022}. Unlike general-purpose systems, they are optimized for dedicated functionalities and often operate under real-time requirements. These characteristics make them well suited for \textit{Edge AI}, where data processing is performed locally at the data source rather than in centralized cloud infrastructures.
\Acp{mcu} are widely used in such systems due to their low power consumption and tight hardware-software integration. To manage concurrent tasks such as sensor acquisition and inference, real-time operating systems (RTOS), e.g., FreeRTOS, are commonly employed. Efficient scheduling and resource management are essential to ensure deterministic behavior and system reliability \cite{liu_scheduling_1973}.

\subsubsection{Sensor Systems and Multimodal Sensing}

Sensor systems provide the interface between the physical and digital domains by capturing environmental and physiological signals \cite{fraden_handbook_2016}. In mobile and dynamic environments, \acp{imu}/\acp{mpu} are particularly relevant for motion analysis, as they enable real-time tracking of movement and orientation \cite{noureldin_fundamentals_2013}. However, such signals are often affected by external disturbances, including vibrations and environmental motion.

To improve robustness, recent approaches increasingly rely on multimodal sensing, combining inertial data with additional modalities such as optical sensing, \ac{hr} monitoring, and environmental context. This fusion enables more reliable inference of human states, especially in scenarios where traditional measurements such as \ac{eeg} or \ac{ecg} are impractical \cite{king_wearable_2021, cook_ambient_2009}.

\subsubsection{TinyML and On-Device Learning}

A key enabler of Edge AI on embedded hardware is \ac{tinyml}, which focuses on deploying machine learning models on resource-constrained \acp{mcu} \cite{ray_review_2022, heydari_tiny_2025}. This requires optimization techniques such as quantization, pruning, and architecture simplification to meet strict memory and latency constraints \cite{han_deep_2016}.

Frameworks such as TensorFlow Lite for Microcontrollers provide support for deploying lightweight models across heterogeneous platforms \cite{david_tensorflow_2021}. Recent research highlights a transition toward more advanced deep learning approaches tailored for embedded environments, along with improved toolchains and deployment workflows \cite{somvanshi_tiny_2025, lamaakal_comprehensive_2025}.

\subsubsection{Sleep and Drowsiness Detection}

Sleep and drowsiness detection has been widely studied in both healthcare and automotive contexts. Traditional approaches often rely on activity-based metrics derived from wearable sensors, which tend to be unreliable in the presence of external motion. More recent work demonstrates the benefits of multimodal approaches, combining wearable, smartphone, and contextual data \cite{hoang_knowledge_2023, martinez_improved_2020, ciman_smartphones_2019}.

In automotive scenarios, driver drowsiness detection systems increasingly integrate physiological signals, motion patterns, and camera-based eye tracking. Vision-based methods using facial landmarks and eye-state classification have shown strong performance in real-time applications \cite{hasan_car_2023, pokhrel_drowsy_2023}. Complementary studies further demonstrate the predictive value of physiological signals such as \ac{hr} for cognitive performance and fatigue assessment \cite{arai_--wild_2024, ajemian_knight_2024}.

\subsubsection{Signal Processing in Dynamic Environments}

A central challenge in mobile scenarios is the presence of noise and artifacts in sensor data caused by external motion. \Ac{dsp} techniques such as adaptive filtering, sliding window analysis, and frequency-domain transformations are commonly applied to separate relevant signals from disturbances \cite{oppenheim_discrete-time_2013}.

Recent studies show that adaptive filtering methods can effectively improve sleep detection performance under noisy conditions \cite{chen_comparative_2024, kim_dynamic_2015}. These preprocessing steps are critical for robust feature extraction and directly impact the performance of downstream machine learning models \cite{sharma_automated_2022}.

\section{Requirements Analysis and System Design}
\label{ch:chapter03}
This chapter defines the application framework and derives suitable use cases. These serve as the foundation for the requirements imposed on the system and its associated Edge AI models.

\subsection{Definition of Application Scenarios}
\label{s:betriebsrahmen}

\subsubsection{Environments with Unsteady Conditions}

\begin{figure}[ht!]
    \centering
    \includegraphics[width=0.22\textwidth]{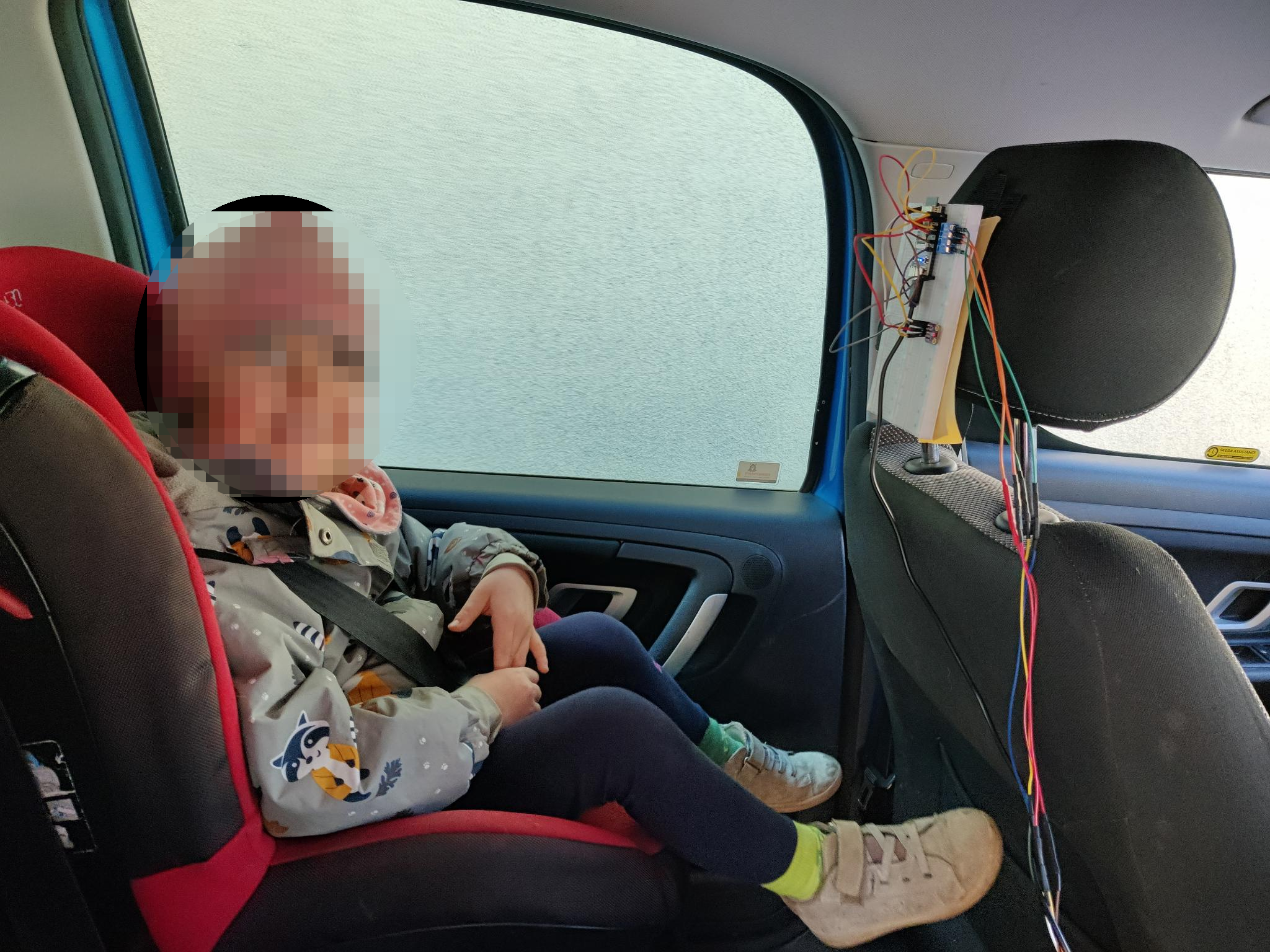}
    \includegraphics[width=0.22\textwidth]{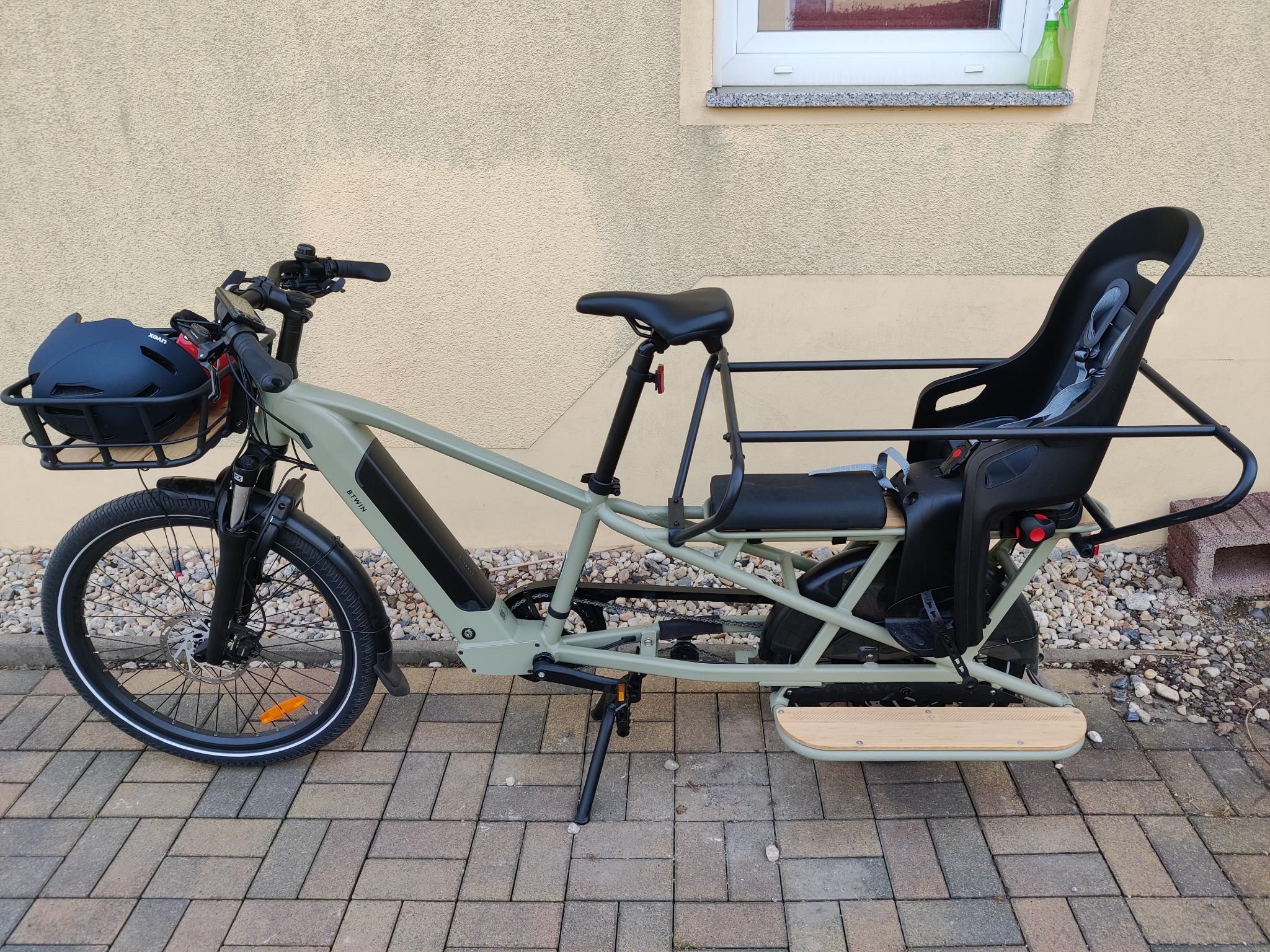}
    \caption{Examples of dynamic, unsteady environments: car (left) and cargo bike (right) measurement setups.}
    \label{fig:betriesrahmen_combined}
\end{figure}

In this work, an \textit{unsteady environment} for sensor-based sleep state detection is defined as a mobile or stimulation-rich setting in which motion, vibration, and varying light or noise conditions render biosignals non-stationary, thereby complicating reliable classification. Typical examples include moving vehicles such as cars or bikes (see Fig. \ref{fig:betriesrahmen_combined}), where external influences continuously introduce artifacts into the sensor data.

Such environments are characterized by strong temporal fluctuations in signal quality, leading to reduced signal-to-noise ratios and unstable feature representations \cite{imtiaz_systematic_2021}. In practice, this manifests as unpredictable artifacts, drifting baselines, and brief or unstable transitions between sleep and wake states. As a result, conventional sleep classification approaches, designed for stationary environments, show limited performance and require adaptation to these conditions.

\subsubsection{Experimental Environments and Application Scenarios}


The experimental evaluation considers two representative unsteady environments that combine practical relevance with distinct signal disturbances. The developed methods are transferable to similar domains but are validated in the following settings:

\paragraph{Moving Car:}
The car represents a moderately unsteady environment with low- to medium-frequency vibrations, intermittent disturbances (e.g., braking, road irregularities), and varying light and noise conditions. It serves as a practical scenario for both sleep monitoring and fatigue detection. Experiments were conducted using two passenger vehicles (see Fig. \ref{fig:betriesrahmen_combined}, left).

\paragraph{Bicycle Trailer / Cargo Bike:}
This setup reflects a highly unsteady environment with strong vibrations, high-frequency shocks, and significant relative motion between subject and sensors. It acts as a stress test for system robustness under real-world conditions. Experiments were conducted using a BTWIN R500E cargo bike (see Fig. \ref{fig:betriesrahmen_combined}, right).

These environments motivate two primary application scenarios. First, sleep monitoring for children (1--6 years) during transit enables non-invasive detection of sleep and wake phases, supporting daily routine planning and longitudinal sleep analysis. Second, fatigue detection for adult drivers focuses on identifying reduced alertness and microsleep events to enable early warning and intervention strategies. Together, these scenarios highlight the potential of embedded, sensor-based systems for real-world sleep and fatigue monitoring under unsteady conditions.

\subsection{Criteria for Sleep State Detection}
\label{s:schlaf}

For sensor-based detection of sleep states, it is essential to consider its various types of stages. Restful sleep is created through the cyclical interaction of different sleep phases, which fulfill distinct functions for physical regeneration, memory formation, and emotional processing. The composition of this sleep architecture depends heavily on age as well as individual characteristics like health status, lifestyle, or genetic influences. Each phase can be distinguished by characteristic features, which provide the bridge to technical analysis \cite{guion_sleep_2011}. Disruptions in this sleep architecture, which can lead to various sleep problems, must be considered as potential sources of error in sensor-based acquisition.

\subsubsection{Sleep Phases and Feature Extraction}

There are several scientifically documented sleep states. The central stages are the wake state, non-\ac{rem} sleep (further divided into light and deep sleep), and \ac{rem} sleep \cite{noauthor_aasm_nodate}. Each state is recognizable via specific physiological features, documented through \ac{eeg}, \ac{ecg}, and other measurement methods \cite{rechtschaffen_manual_1968}:

\begin{itemize}
    \item Wake state: Consciousness, reactive motor activity, normal \ac{hr} and respiration.
    \item Sleep Onset (Stage 1, N1): Transition from wakefulness to sleep, slow eye movements, onset of muscle relaxation, decreasing alpha waves in \ac{eeg}.
    \item Non-\ac{rem} Sleep:
    \begin{itemize}
        \item Light Sleep (Stage 2, N2): Sleep spindles and K-complexes in \ac{eeg}, increasingly relaxed musculature, no eye movements, steady \ac{hr}.
        \item Deep Sleep (Stage 3, N3): Delta waves in \ac{eeg}, lowest \ac{hr}, slow breathing, minimal body movement.
    \end{itemize}
    \item \ac{rem} Sleep: Very low muscle tone, increased \ac{hr} and blood pressure, irregular and accelerated breathing.
\end{itemize}


\subsubsection{Duration of Sleep Phases}

This work focuses on short sleep episodes in unsteady environments rather than nocturnal sleep. Such episodes are typically referred to as \textit{power naps} (10--30\,min), primarily involving N1 and N2 stages, while deeper sleep (N3) may occur in longer durations (30--90\,min). \ac{rem} sleep is usually reached only in extended sleep periods.

Sleep architecture strongly depends on age. Infants and young children exhibit shorter sleep cycles and higher \ac{rem} proportions (up to 50\%), whereas adults show longer cycles (90--120\,min) with reduced REM and deep sleep shares \cite{schlieber_role_2021}. For the considered applications, N1--N3 stages are most relevant; however, age-dependent variations imply that \ac{rem}-related characteristics may still be present even in short sleep episodes.

\section{Technical Implementation and System Architecture}
\label{sec:implementation}

The proposed sleep-detection system was designed as an autonomous, battery-powered embedded platform that performs sensing, preprocessing, and inference locally. This on-device processing enables mobile operation without permanent external connectivity and reduces privacy risks by avoiding the transmission of raw sensor data.

\subsection{Hardware Platform}
\label{subsec:hardware-platform}

The system was built around an \textit{ESP32-S3} \ac{mcu}, selected for its dual-core architecture, support for external PSRAM, and suitability for embedded Edge AI workloads. In contrast to classic ESP32 variants, the ESP32-S3 offers higher computational capability and better memory support for tasks such as image acquisition, feature extraction, and lightweight neural inference.

For mobile operation, the platform was powered by a \textit{18650 battery shield} with two series-connected lithium-ion cells. The resulting supply voltage is regulated to the levels required by the ESP32-S3, while integrated charging and protection circuitry supports safe battery operation. Depending on workload and duty cycle, the runtime can extend over several hours; continuous operation with active wireless communication or camera usage significantly increases current consumption.

\subsection{Sensor Configuration}
\label{subsec:sensor-configuration}

The implemented prototype used two sensor modalities: a \ac{mpu} (\textit{MPU6050}, 100\,Hz) and a camera module (\textit{OV3660}, 0.2\,Hz). The inertial sensor was used to capture head and body motion and the camera supported visual observation of relevant pose information. Additional sensing modalities were evaluated but found to be unsuitable for robust sleep-state inference in unsteady environments. An optical \ac{hr} sensor enabled physiological pulse measurement but suffered from instability under motion. Environmental sensors, including a microphone, temperature sensor, and light sensor, were also considered; however, their signals are highly affected by ambient noise and varying external conditions, limiting their reliability and discriminative power for sleep-state detection.

\begin{figure}[ht!]
\centering
\includegraphics[width=.35\textwidth]{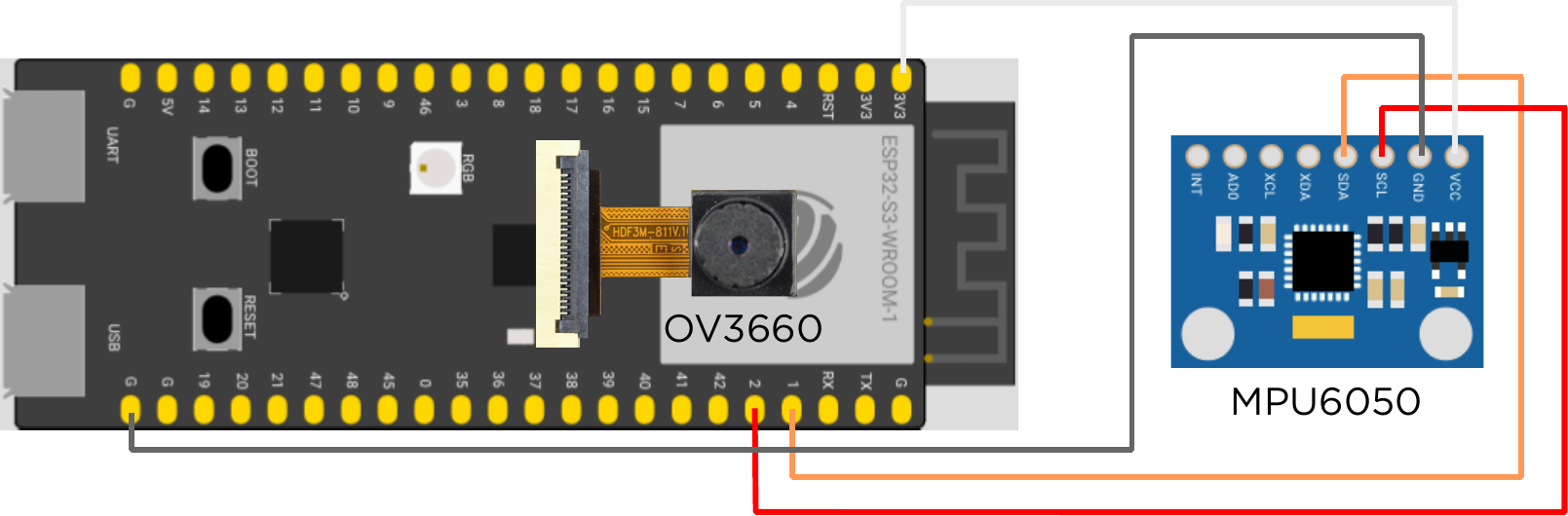}
\caption{Hardware platform with ESP32-S3, OV3660 and MPU6050.}
\label{fig:platform}
\end{figure}

As shown in Fig. \ref{fig:platform}, the \ac{mpu} was connected via I\(^2\)C and provided tri-axial acceleration and gyroscope measurements. These signals were used to derive posture- and motion-related features that support sleep-awake classification. The camera was interfaced through a dedicated camera connector and configured for low-frequency image acquisition to reduce memory usage and computational load.

\subsection{Task Parallelization}
\label{subsec:parallelization}

The software structure followed a modular design in which each sensor and processing component is encapsulated in a dedicated class. This approach improved maintainability and allows individual subsystems to be extended or replaced independently. Core pinning, queue-based communication, and mutex-protected access together formed the basis for robust real-time operation on the embedded platform.

\begin{figure}[h!]
    \centering
    \includegraphics[width=1.0\linewidth]{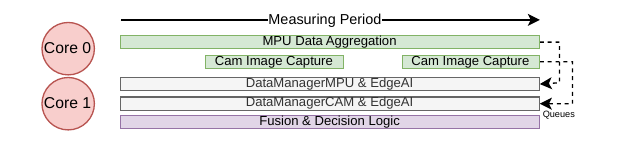}
    \caption{Parallelization of sensor tasks (Core 0: continuous \ac{mpu} measurements and event-driven image capture by the camera) and data analysis (Core 1: DataManager and decision logic), each handled by a dedicated processor core.}
    \label{fig:messablauf}
\end{figure}

The ESP32-S3 dual-core architecture was exploited using FreeRTOS tasks pinned to specific cores. This separation allows time-critical sensor acquisition to run independently from computationally heavier data processing and inference tasks. A dedicated sensing core handled periodic sensor sampling, while the second core executed data management, storage, and decision logic (see Fig. \ref{fig:messablauf}).

\begin{table}[ht!]
\caption{Task allocation across the ESP32-S3 cores.}
\label{tab:task-allocation}
\begin{tabular}{clp{4cm}}
\hline
Core & Task & Function \\
\hline
Core 0 & MPU task & Continuous inertial data acquisition \\
Core 0 & Camera task & Event-driven image capture \\
Core 1 & Data manager tasks & Data aggregation, preprocessing, and inference (Sec. \ref{sec:edge-ai}) \\
Core 1 & SD manager & Storage of logged data (Sec. \ref{subsec:data-buffering})\\
Core 1 & Sleep manager & Decision logic \& fusion (Sec. \ref{ss:fusionlogic}) \\
\hline
\end{tabular}
\end{table}

This architecture reduced scheduling conflicts and improved determinism during sensor acquisition. The camera task could be suspended when image capture is not required, which further lowered resource usage. Task coordination and runtime control were implemented in a central task manager that initializes tasks, assigns cores, and controls their execution state.

\subsection{Data Buffering and Synchronization}
\label{subsec:data-buffering}

\begin{figure}[ht!]
    \centering
    \includegraphics[width=1.0\linewidth]{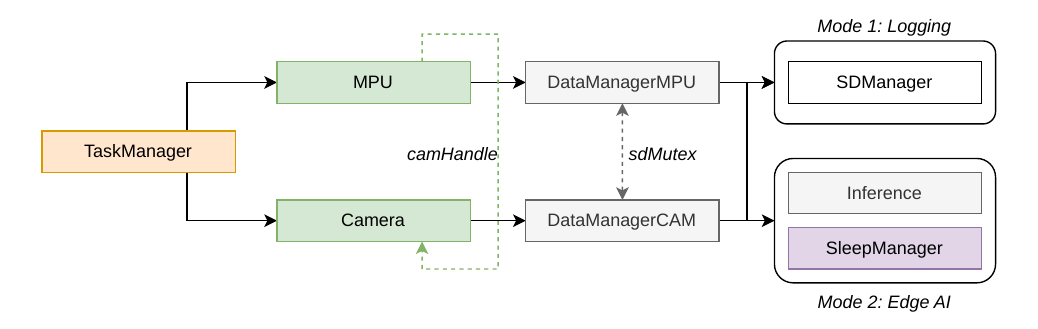}
    \caption{System operating modes: Mode 1 for data storage, Mode 2 for inference of Edge AI applications.}
    \label{fig:architektur}
\end{figure}

Sensor data were transferred between acquisition tasks and processing tasks using FreeRTOS queues. This producer-consumer structure buffered temporary load peaks and supported loss-free communication between tasks. Each data sample was timestamped to enable later synchronization and fusion across modalities. The system could run in two separate modes: logging mode (mode 1) and Edge AI mode (mode 2). In mode 1, sensor batches were written to SD storage to reduce write operations and improve storage efficiency; in mode 2, the system operated without persistent logging and focused on inference. The logging process to SD was synchronized via a global mutex (\textit{sdMutex}) to avoid collisions (see Fig. \ref{fig:architektur}).


\section{Data \& Model Preparation}

\subsection{Physical Setup and Sensor Positioning}
\label{subsec:physical-setup}

The sensor configuration ensured accurate capture of head motion and facial features across automotive (front/rear seats) and cargo-bike scenarios. The MPU6050 inertial sensor was mounted rigidly to the subject's head, with coordinate system aligned as: $x$-axis lateral (right), $y$-axis anterior (forward), $z$-axis cranial (upward). This right-handed body-fixed frame enabled unambiguous tracking of head rotations and translations. Initial calibration in rest state compensated positioning offsets.

The camera positioning optimized face capture within a potential \ac{roi}-cropping from QVGA images (see Sec. \ref{sss:cam_data}). Using the pinhole camera model~\cite{szeliski_computer_2010} with an assumed face width of 16 cm, Table~\ref{tab:mounting-positions} summarizes operational distances across scenarios.

\begin{table}[ht!]
\centering
\caption{Sensor-to-face distances $d$ for 16 cm face width assumption.}
\label{tab:mounting-positions}
\begin{tabular}{ccc}
\hline
Scenario & Mounting Position & $d$ \\
\hline
Car (driver) & Windshield phone holder & 40 cm \\
Car (rear seat) & Headrest (see Fig. \ref{fig:montage}) & 41 cm \\
Cargo-bike (child seat) & Driver seat post & 38 cm \\
\hline
\end{tabular}
\end{table}

\begin{figure}[ht]
\centering
\includegraphics[height=3.5cm]{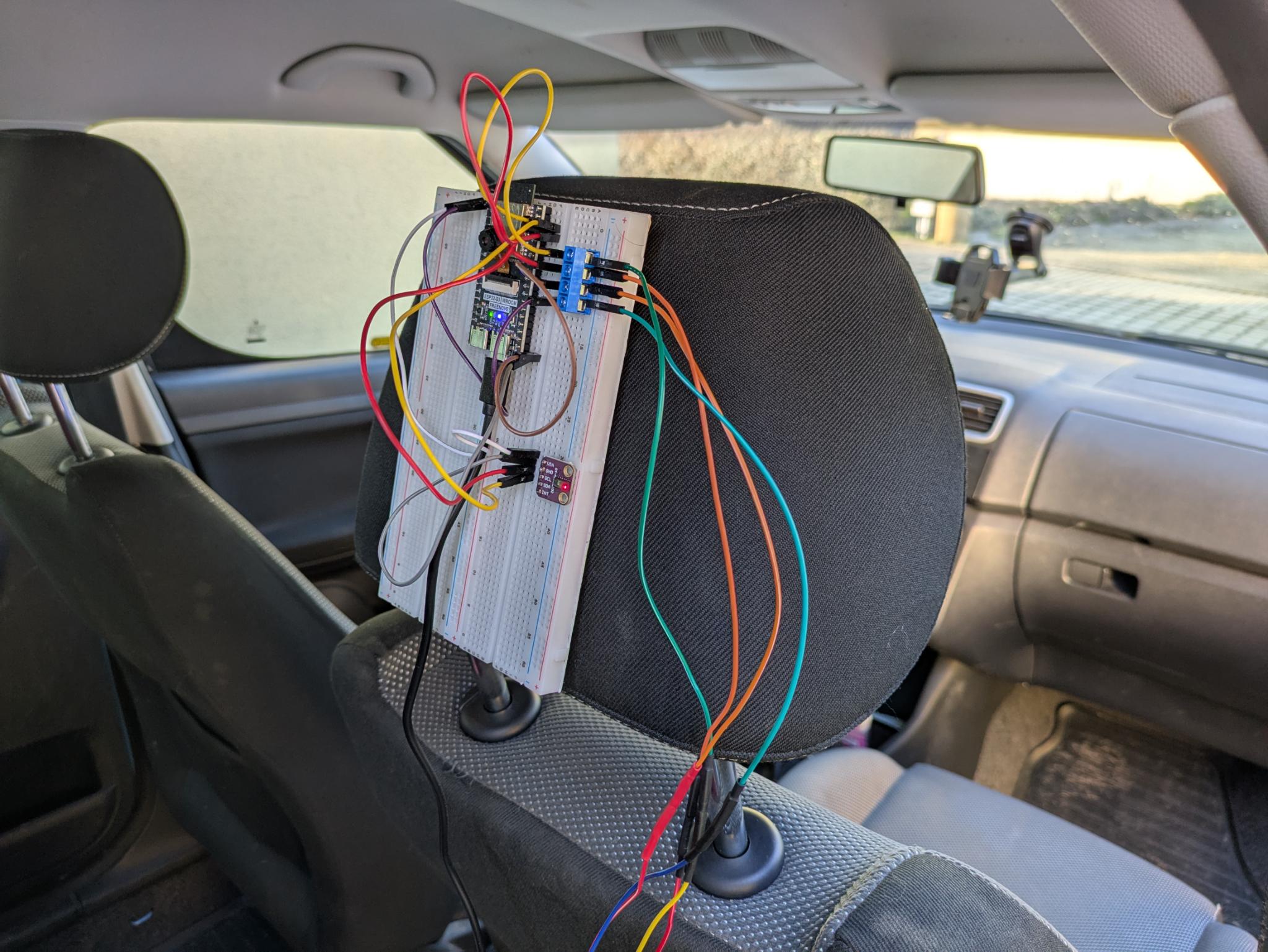}
\caption{Hardware mounting with 3M Velcro on headrest (rear seat).}
\label{fig:montage}
\end{figure}

Mechanical mounting used separate breadboards for MPU6050 and ESP32 with 60 cm jumper cables (fixed via 4-pin screw terminals, heat-shrink reinforced). I$^2$C operated at 50 kHz to accommodate cable length without signal degradation. All components  were fixed via 3M Velcro for robust, removable attachment (Fig.~\ref{fig:montage}). The ESP32-S3 camera was mounted in phone holders (driver), headrests (rear) or seat post (bike); the \ac{mpu} was attached on the participant's hat.

In terms of connectivity, a wireless \ac{ble} 5.0 stack was initialized in the TaskManager to share inference results in a mobile setting. The TaskManager initialized a GATT service with Notify characteristic, while the SleepManager fed inference results. The exemplary Android app BLE Radar\footnote{\url{https://github.com/BLE-Research-Group/MetaRadar}} visualized the sleep classifications results.



\subsection{Data Aggregation and Preparation}
\label{subsec:data-aggregation}

Data collection targeted binary classification tasks: (1) head motion vs. rest from MPU6050 inertial signals (Sec. \ref{sss:mpudata}), and (2) sleep poses from OV3660 camera images (Sec. \ref{sss:cam_data}). The measurements spanned automotive (front/rear seats) and cargo-bike scenarios to capture environmental variability.

\subsubsection{Inertial Dataset} 
\label{sss:mpudata}
\begin{figure}[ht!]
    \centering
    \includegraphics[width=.9\linewidth]{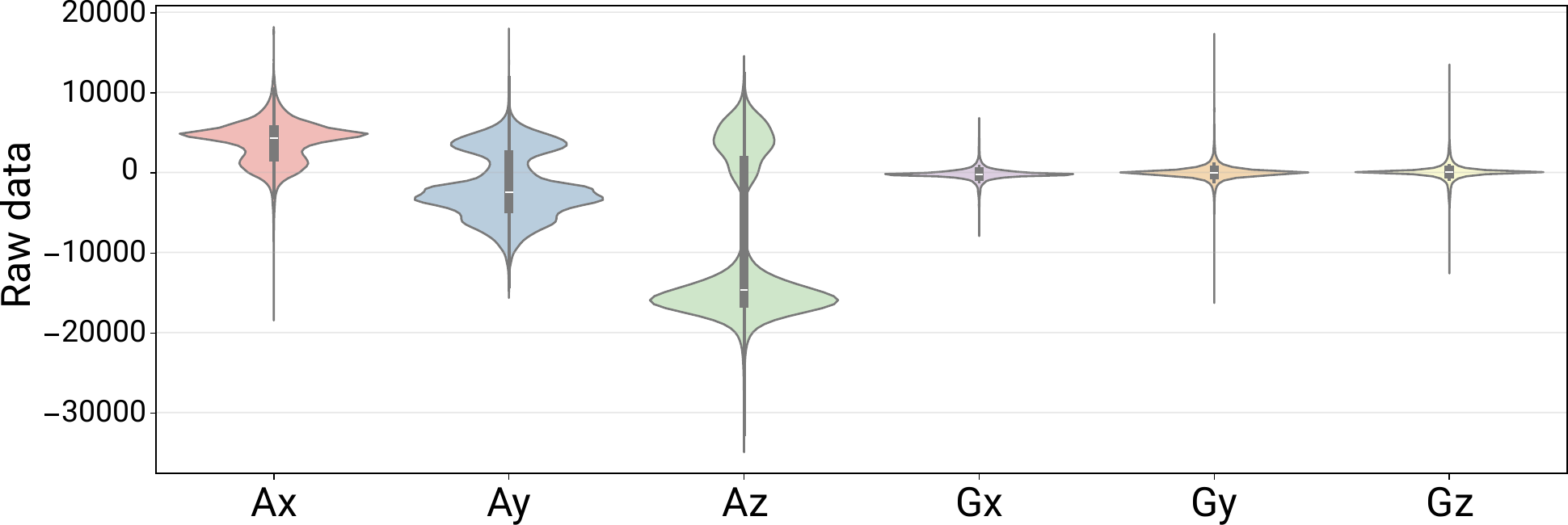}
    \includegraphics[width=.9\linewidth]{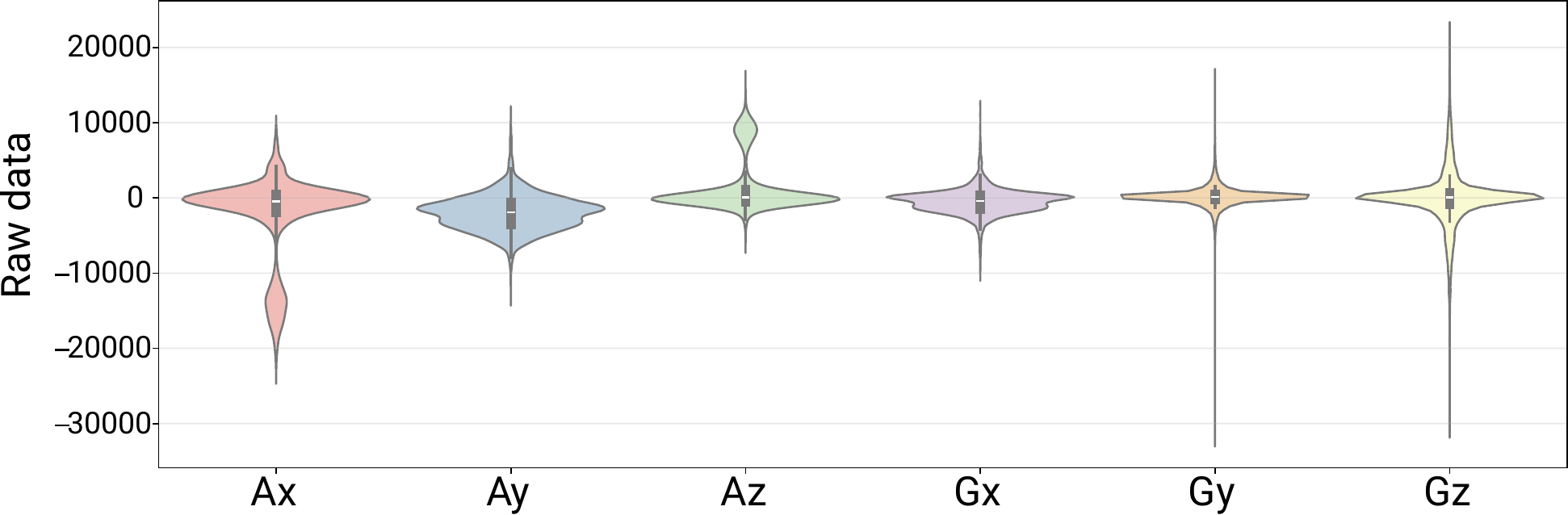}
    \caption{Feature values for \acs{mpu} raw data in the car without head movements (top) and with head movements (bottom).}
    \label{fig:onlycar_violin}
\end{figure}

124,343 samples at 100\,Hz from two adults and two children were collected, comprising 76,686 \textit{no intentional motion} and 47,657 \textit{with motion} (head turns, nods). Automotive data dominated (68,586/37,657), with cargo-bike data at 8,100/10,000. The sliding windows of 100 samples (1\,s) yielded 5,458 segments (3,406 motion, 2,052 rest) after edge trimming. Feature values and statistical metrics of the aggregated data are presented in Fig. \ref{fig:onlycar_violin} (car) and Tab. \ref{tab:gyro_stats_full}.

\begin{table}[ht!]
\centering
\caption{Statistical metrics of gyroscope raw data (aggregated datasets).}
\label{tab:gyro_stats_full}
\begin{tabular}{c|c|c|c|c|c}
\hline
\textbf{Vehicle / Head} & \textbf{Axis} & \textbf{$\mu$} & \textbf{$\sigma$} & \textbf{Min} & \textbf{Max} \\
\hline
Car / idle  & Gx & -223.46 & 774.47  & -7769  & 6616  \\
            & Gy & -31.47  & 991.56  & -16080 & 17087 \\
            & Gz & 19.05   & 896.42  & -12412 & 13264 \\
\hline
Car / moving & Gx & -496.07 & 1608.26 & -10577 & 12455 \\
            & Gy & 141.03  & 1269.83 & \cellcolor{lightgray}-32699 & 16782 \\
            & Gz & -38.81  & 3558.47 & -30876 & 22378 \\
\hline
Bike / idle & Gx & -55.15  & 1114.14 & -6841  & 7967  \\
            & Gy & -70.78  & 752.21  & -3834  & 2923  \\
            & Gz & 140.95  & 933.05  & -7673  & 5701  \\
\hline
Bike / moving & Gx & -42.37  & 4489.70  & -27390 & 26369 \\
             & Gy & 59.43   & 4704.62  & -24971 & 21147 \\
             & Gz & 328.72  & 10553.70 & -32513 & \cellcolor{lightgray}32461 \\
\hline
\end{tabular}
\end{table}




\subsubsection{Camera Dataset} 
\label{sss:cam_data}
In total, 1{,}178 QVGA grayscale images were collected from four subjects (two adults and two children) in the car. The dataset comprises 175 awake and 163 sleep samples from the front seat, and 341 awake and 499 sleep samples from the rear seat. Data acquisition was performed during urban driving scenarios, incorporating variations such as overexposure and head pose changes to improve robustness. 

\begin{figure}[ht!]
    \centering
    \includegraphics[width=0.4\linewidth]{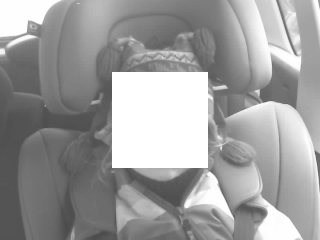}\\
    \includegraphics[width=0.4\linewidth]{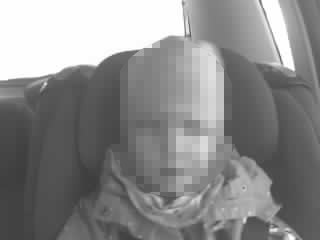}
    \includegraphics[width=0.4\linewidth]{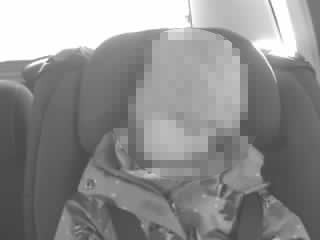}
    \caption{\acs{roi}-Crop ($96 \times 96$ px) of a QVGA image in grayscale mode (top). Awake state with full set of facial features (bottom left) and sleep state with pose change and missing facial features (bottom right).}
    \label{fig:cam_crop}
\end{figure}

The visual approach implicitly leveraged pose estimation by focusing on the presence of facial features within a fixed \ac{roi} (Fig. \ref{fig:cam_crop}, top). The \acp{roi} were defined around the expected face position (e.g., in a fixed child seat) and cropped with a size of $96 \times 96$ pixels, enabling a consistent spatial reference. During wakefulness, characteristic facial features were reliably captured within this region (Fig. \ref{fig:cam_crop}, bottom left). In contrast, sleep-related head poses, such as a forward or lateral tilt, often lead to partial or complete disappearance of these features (Fig. \ref{fig:cam_crop}, bottom right).

As a result, the Edge AI model learned the presence or absence of facial structures as an indirect indicator of head pose. This made the approach particularly suitable for constrained setups with limited subject movement, where the \acp{roi} remained spatially stable and could serve as a proxy for pose estimation. 




\section{Edge AI Implementation}
\label{sec:edge-ai}

This section presents the Edge AI pipeline for sleep detection on the ESP32-S3 \ac{mcu}. Separate models processed inertial data for head motion classification and camera images for pose analysis. A sequential decision framework activated resource-intensive vision processing only after sustained motionlessness, balancing computational efficiency with detection accuracy in mobile scenarios.

\subsection{Training Pipeline and Optimization}
\label{subsec:training}

TensorFlow-based \ac{tinyml} models were developed in Python 3.13.5 using virtual environments, leveraging \textit{Keras 3.13.2, NumPy 2.4.2, pandas 3.0.1, scikit-learn 1.8.0}, and \textit{TensorFlow 2.20.0}~\cite{abadi_tensorflow_2016}. Both \ac{mpu} and camera models served as proof-of-concept implementations demonstrating the deployment principle and remain interchangeable with alternative architectures. The training and deployment architecture comprised the following steps:

\begin{enumerate}
    \item \textit{Preprocessing:} Sliding windows (\ac{mpu}), \ac{roi} (camera), StandardScaler normalization
    \item \textit{Splitting:} Stratified 80/10/10 train/valid/test
    \item \textit{Training:} GPU-accelerated, early stopping option
    \item \textit{Quantization:} Post-training \texttt{int8} quantization
    \item \textit{Export:} Conversion of TensorFlow Lite model into a C-compatible byte array representation for deployment 
\end{enumerate}

Feature standardization was performed using \textit{StandardScaler} from the \textit{scikit-learn} library. To ensure consistent behavior on the ESP32 platform, the corresponding scaling parameters ($\mu, \sigma$) were precomputed and exported to the \ac{mcu}. These parameters were integrated into the embedded implementation via a dedicated header file. On ESP32-S3, the framework TensorFlow Lite Micro \cite{david_tensorflow_2021} was used.

\subsection{Motion Detection Model (MPU6050)}
\label{subsec:mpu-model}

A \ac{mlp} classified 1\,s motion windows from 700-dimensional input vectors (100$\times$7 features: triaxial accelerometer/gyroscope + rotational energy norm):
\begin{equation}
\label{eq:rotation}
E_i = \sqrt{Gx_i^2 + Gy_i^2 + Gz_i^2}
\end{equation}
An initial calibration subtracted rest-state offsets (100-sample means per axis). No explicit filtering was applied due to computational costs. The model learned directly from raw signals, with the rotational energy norm (Eq. \ref{eq:rotation}) as additional feature.

\subsubsection{Architecture} The employed \ac{mlp} consisted of an input layer with 700 neurons, followed by two fully connected hidden layers comprising 20 and 10 neurons, respectively. Both hidden layers used ReLU activation and a dropout rate of 0.3 to mitigate overfitting. The network terminated in a sigmoid-activated output neuron for binary classification. 

\subsubsection{Performance} 
\begin{figure}[ht!]
    \centering
    \includegraphics[width=1.0\linewidth]{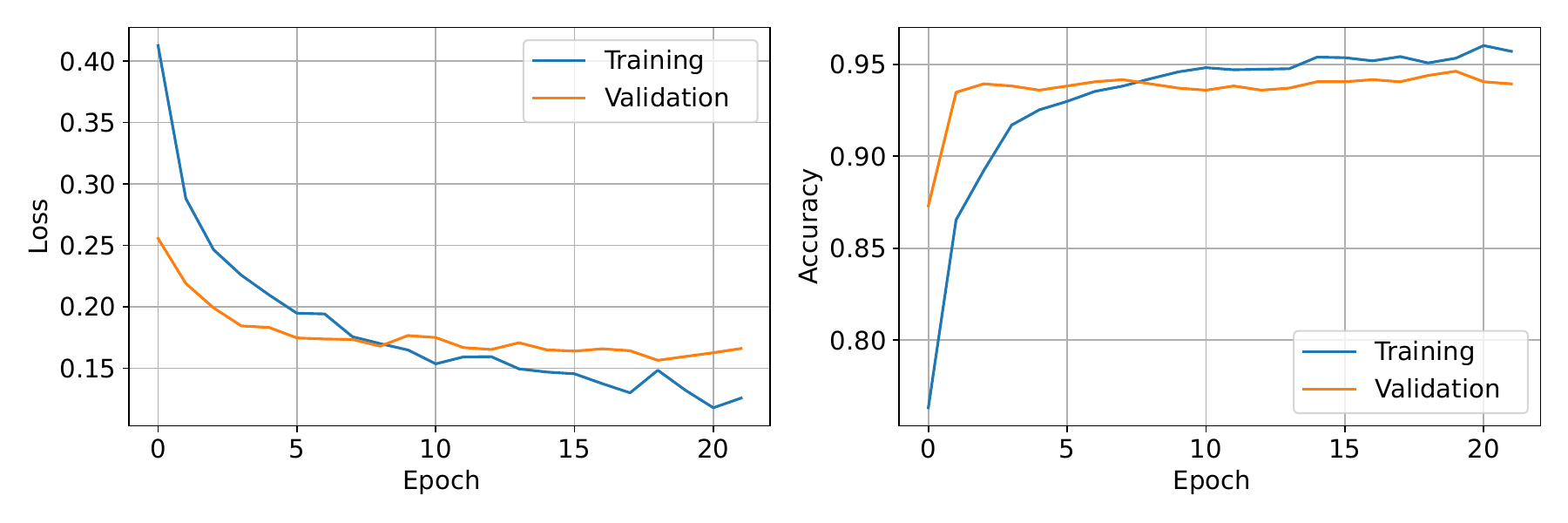}
    \caption{Trend in loss and accuracy of the Edge AI model for the \ac{mpu} following early stopping at 22 episodes.}
    \label{fig:mpu_ai_results}
\end{figure}

Training was conducted using the Adam optimizer \cite{kingma_adam_2017} with a learning rate of $\eta = 5 \times 10^{-4}$ and binary cross-entropy loss. Early stopping was applied, with convergence observed at epoch 22 (cf. Fig.~\ref{fig:mpu_ai_results}).

The proposed model achieved a test accuracy of 96.5\% across 1{,}092 evaluation windows. The corresponding confusion matrix is presented in Table~\ref{tab:mpu-confusion} with motion covering movement of the head - independent of the environment.

\begin{table}[h]
\centering
\caption{Confusion matrix of the MPU6050-based motion classification.}
\label{tab:mpu-confusion}
\begin{tabular}{lcc}
\hline
\multicolumn{1}{c}{True$\backslash$Pred} & No Motion & Motion \\
\hline
No Motion & \cellcolor{lime}667 & \cellcolor{pink}14 \\
Motion    & \cellcolor{pink}24  & \cellcolor{lime}387 \\
\hline
\end{tabular}
\end{table}

The model yielded F1-scores of 0.97 for the no-motion class and 0.95 for the motion class, with an overall ROC-AUC of 0.99.

\subsection{Visual Classification Model (OV3660)}
\label{subsec:cam-model}

\subsubsection{Architecture} A compact \ac{cnn} was employed to process $96 \times 96$ grayscale face \acp{roi}. To improve generalization, data augmentation techniques were applied during training, including horizontal flipping, random rotations within $\pm 5^\circ$, and zoom variations of up to 10\%.

The network architecture consisted of three convolutional blocks with 8, 16, and 24 filters, respectively. Each block used $3 \times 3$ kernels, followed by ReLU activation and $2 \times 2$ max-pooling. The convolutional backbone was followed by a global average pooling layer and a sigmoid-activated output neuron for binary classification. 

\subsubsection{Performance} 
\begin{figure}[ht!]
    \centering
    \includegraphics[width=1.0\linewidth]{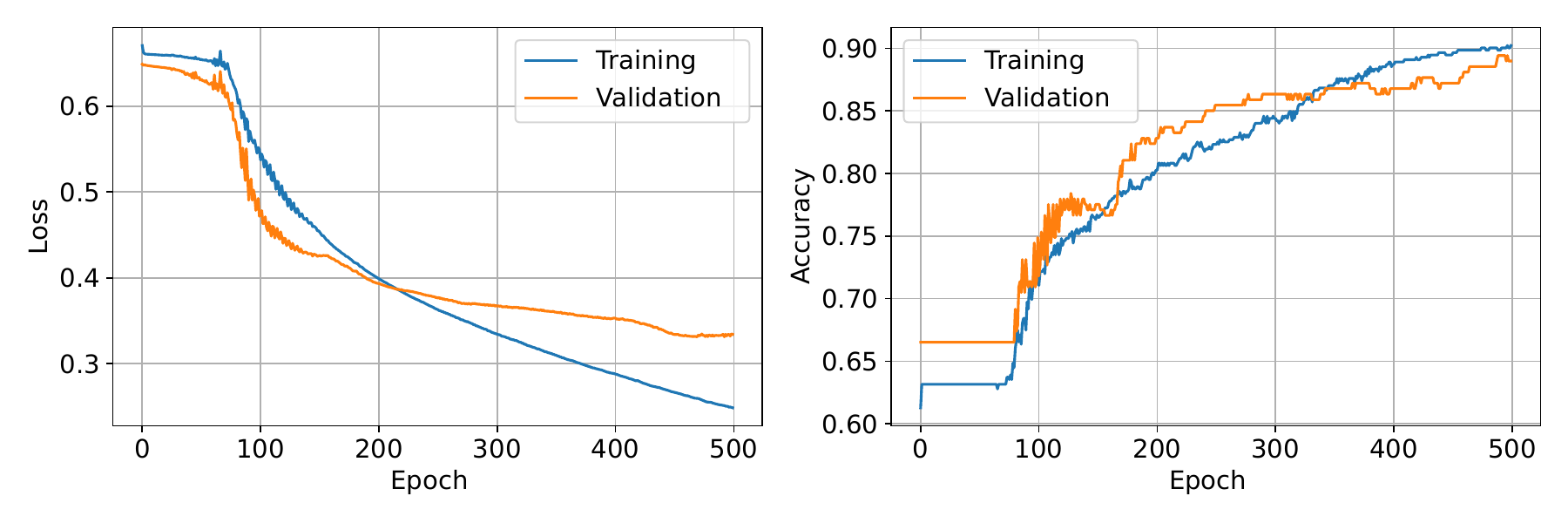}
    \caption{Trend in accuracy and loss of the Edge AI model for the camera following early stopping at 500 episodes.}
    \label{fig:cam_ai_results}
\end{figure}

Training was performed also using the Adam optimizer and binary cross-entropy loss. Early stopping was applied to prevent overfitting, with convergence observed after 500 epochs (see Fig. \ref{fig:cam_ai_results}).

The proposed model achieved a test accuracy of 89\% on a dataset comprising 227 images. The corresponding confusion matrix is presented in Table~\ref{tab:cam-confusion}.

\begin{table}[h]
\centering
\caption{Confusion matrix of the camera-based pose classification.}
\label{tab:cam-confusion}
\begin{tabular}{lcc}
\hline
\multicolumn{1}{c}{True$\backslash$Pred} & Sleep & Awake \\
\hline
Sleep & \cellcolor{lime}64 & \cellcolor{pink}12 \\
Awake   & \cellcolor{pink}14 & \cellcolor{lime}137 \\
\hline
\end{tabular}
\end{table}

The model achieved a macro-averaged F1-score of 0.87, with class-wise F1-scores of 0.83 for the sleep class and 0.91 for the awake class. 


\subsection{Sensor Fusion \& Decision Logic}
\label{ss:fusionlogic}

\begin{figure}[ht!]
\centering
\includegraphics[width=.5\textwidth]{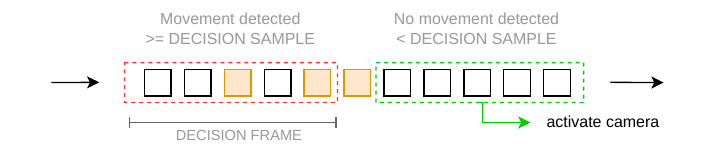}
\caption{Sensor fusion and decision logic with \colorsquare{white} as  \ac{mpu} inference class 0 (no movement) and \colorsquare{peachlight} as class 1 (movement).}
\label{fig:logic}
\end{figure}

The system employed a hierarchical sensor fusion strategy to minimize computational load while maintaining robust decision-making. Initially, inertial data from the \ac{mpu} was continuously analyzed to detect motion. Only in the absence of movement in a given amount of time (\texttt{DECISION\_FRAME}), the camera-based \ac{cnn} was activated to perform visual classification. If the model predicted an awake state (class 0), motion monitoring continued. Otherwise, a confidence-based threshold was applied to validate the detection (\texttt{DECISION\_SAMPLE}). Optionally, an additional verification stage could be incorporated before final classification. This event-driven activation (see Fig. \ref{fig:logic}) of sensing modalities ensured efficient resource utilization while preserving reliable sleep state inference.


\subsection{Deployment \& System Validation}
\label{subsec:validation}

The \ac{mpu} model comprised 14{,}241 trainable parameters, corresponding to a size of $57.86\,\text{KB}$ in \texttt{float32} representation. After \texttt{int8} quantization, the total model size was reduced to approximately \underline{$17\,\text{KB}$}. For deployment on the target platform, the TensorArena was allocated with $16\,\text{KB}$, matching the memory requirements of the TensorFlow Lite model. The inference time on the ESP32-S3 averaged \underline{2.39 ms}.

The camera model comprised 5{,}145 trainable parameters with a total size of \underline{$11\,\text{KB}$} after \texttt{int8} quantization. Due to increased runtime memory requirements, the TensorArena size was set to $512\,\text{KB}$ to ensure stable execution without memory overflow. The inference time on the ESP32-S3 averaged \underline{2.7 s}.

A final system validation focused on verifying the correct interaction of all components after deployment on the ESP32-S3. In particular, the decision logic (Sec. \ref{ss:fusionlogic}) and the parallelization of inertial and vision-based classification were assessed under representative real-world scenarios.

A set of functional tests (30\,s scenarios, $n=10$ per condition, \texttt{DECISION\_FRAME=3}, \texttt{DECISION\_SAMPLE=5000}) was conducted with two participants, simulating sleep and wake conditions through controlled head movements and pose changes. The evaluation considered combinations of vehicle motion and head activity in both bicycle and car settings. Results based on the given settings indicate that wake states were reliably detected when head movement is present, achieving consistent performance across all scenarios ($10/10$). The \ac{mpu}-based motion model demonstrated strong reliability, enabling clear separation between motion and rest even in non-stationary environments. Camera activation was consistently triggered upon detected inactivity, confirming correct event-driven operation. 

The vision-based component demonstrated reliable performance when leveraging implicit pose estimation via the defined \ac{roi}. In particular, sleep-related head poses, such as a forward or lateral tilt, were robustly detected due to the resulting absence of facial features within the \ac{roi} ($9/10$ car, $8/10$ bike). However, this approach inherently depended on pronounced head movements and therefore limits applicability to scenarios where such poses occur. Subtle indicators, such as eye closure without significant pose change, were less reliably captured, reducing generalization across different sleep behaviors.



Overall, the results confirmed that the proposed fusion logic operates as intended and enables reliable wake detection, while sleep classification - particularly based on visual cues - remains an area for further improvement. A statistically significant evaluation requires a larger and more diverse dataset and is left for future work.



\section{Conclusion \& Outlook}

This work demonstrates that binary sleep-wake detection can be inferred on resource-constrained embedded platforms using Edge AI methods. The proposed system, implemented on an ESP32-S3, integrates heterogeneous sensors and lightweight machine learning models to enable robust operation in dynamic environments while maintaining energy efficiency suitable for long-term deployment.

Experimental results showed that head motion analysis using inertial data provides a reliable indicator for distinguishing rest and activity, even under non-stationary conditions. Compact models such as the presented \ac{mlp} achieved high performance with minimal resource requirements. In addition, the use of a lightweight \ac{cnn} for visual input is feasible on \ac{mcu} and enables complementary pose-based analysis. However, limitations remain in the robustness due to the necessity of pose changes during sleep for this approach.

Future work should focus on improving generalization through larger and more diverse datasets, as well as enhancing system usability through wireless architectures and integrated user interfaces. Furthermore, extending the visual models toward more robust feature representations may enable finer-grained sleep state classification and improved detection of fatigue-related conditions.

\newpage

\bibliographystyle{IEEEtran}
\bibliography{bibliography}

\input{glossary}


\end{document}

%% file: glossary.tex
\begin{acronym}[tflm]
\acro{adc}[ADC]{analog-to-digital converter}
\acro{alc}{ambient light cancellation}
\acro{api}[API]{application programming interface}
\acro{ble}[BLE]{Bluetooth Low Energy}
\acro{bpm}{beats per minute}
\acro{ci}[CI]{continuous integration}
\acro{cnn}[CNN]{convolutional neural network}
\acro{dl}[DL]{deep learning}
\acro{dma}{direct memory access}
\acro{dvfs}{dynamic voltage and frequency scaling}
\acro{dry}{don't repeat yourself}
\acro{dsp}[DSP]{digital signal processing}
\acro{edf}{earliest deadline first}
\acro{eeg}[EEG]{electroencephalography}
\acro{ecg}[ECG]{electrocardiogram}
\acro{fifo}[FIFO]{first in - first out}
\acro{fov}[FOV]{field of view}
\acro{fpc}[FPC]{flexible printed circuit}
\acro{fpga}[FPGA]{field programmable gate array}
\acro{fps}[FPS]{frames per second}
\acro{gpio}[GPIO]{general-purpose input/output}
\acro{hba}{human behavior analysis}
\acro{hr}[HR]{heart rate}
\acro{hf}[HF]{heart frequency}
\acro{imu}[IMU]{inertial measurement unit}
\acro{iot}[IoT]{Internet of things}
\acro{ai}[AI]{artificial intelligence}
\acro{mcu}[MCU]{microcontroller unit}
\acro{mc}[MC]{microcomputer}
\acro{ml}[ML]{machine learning}
\acro{mlp}[MLP]{multilayer perceptron}
\acro{mpu}[MPU]{motion processing unit}
\acro{ppg}[PPG]{photoplethysmography}
\acro{rem}[REM]{rapid eye movement}
\acro{rms}[RMS]{root mean square}
\acro{roi}[ROI]{region of interest}
\acro{sda}[SDA]{serial data}
\acro{scl}[SCL]{serial clock}
\acro{soc}[SOC]{system-on-chip}
\acro{tinyml}[TinyML]{tiny machine learning}
\acro{tflm}[TFLM]{TensorFlow Lite Micro}
\acro{tinydl}[TinyDL]{tiny deep learning}
\acro{uml}[UML]{unified modeling language}
\end{acronym}